\documentclass[10pt,letterpaper,twocolumn]{article}
\usepackage[]{graphicx}
\usepackage{amsthm}
\usepackage{amsmath}
\usepackage{float} % for figures, from /home/frederik/pictures/creevey/home-improv-2021-06-01/home-improv.tex
\usepackage{abstract}

\author{Frederik Eaton \\
\texttt{\small frederik@gmail.com}
}
\date{September 29, 2022}
\title{Belief propagation generalizes backpropagation}

\newcommand{\partby}[2]{{\partial #1\over \partial #2}}
\newcommand{\ud}{\mathrm{d}}
\newcommand{\dby}[2]{\frac{\ud #1}{\ud #2}}
\newcommand{\abs}[1]{\left| #1 \right|}
\renewcommand{\-}{\setminus}
\newtheorem{thm}{Theorem}
\newtheorem{cor}{Corollary}
\begin{document}
\maketitle

\begin{abstract}
  \noindent
The two most important algorithms in artificial intelligence are
backpropagation and belief propagation. In spite of their importance,
the connection between them is poorly characterized. We show that when
an input to backpropagation is converted into an input to belief
propagation so that (loopy) belief propagation can be run on it, then
the result of belief propagation encodes the result of
backpropagation; thus backpropagation is recovered as a special case
of belief propagation. In other words, we prove for apparently the
first time that belief propagation generalizes backpropagation. Our
analysis is a theoretical contribution, which we motivate with the
expectation that it might reconcile our understandings of each of
these algorithms, and serve as a guide to engineering researchers
seeking to improve the behavior of systems that use one or the other.
\end{abstract}

\section{Introduction}

In this paper we consider a connection between two algorithms which
could be said to have the status of being the two most fundamental
algorithms in the various fields of computer science concerned with
the numerical modeling of real world systems, these fields being
sometimes known as artificial intelligence or machine learning,
sometimes called control theory or statistical modeling or approximate
inference. The two algorithms that form our subject matter are usually
referred to as backpropagation and belief propagation, respectively,
although these are modern terms for concepts that go back several
hundred years in Western thought
\cite{newton1687,leibniz1920,bethe1935sts}.

Back-propagation is another name for the chain rule of differential
calculus \cite{lagrange1797}, applied iteratively to a network of
functions, or in other words to a function of functions of multiple
independent variables and other such functions; the input to
backpropagation may also be known in the field as a ``deep network''
or a ``neural network'' \cite{rumelhart1986learning}.

Belief propagation, by contrast, takes as its input a network of
probability distributions, also called a probabilistic network. Belief
propagation is equivalent to an iterative application of Bayes' rule,
which is the rule for inferring the posterior distribution $P(X|Y)$ of
a random variable $X$ from its prior $P(X)$ and some table of
conditional probabilities $P(Y|X)$ describing the possible
observations of some related variable $Y$: $P(X|Y)$ = ${1 \over Z(Y)}
P(X)P(Y|X)$, often stated as $P(X|Y) = {P(X) P(Y|X) \over P(Y)}$
\cite{bayes1763,laplace1812}. Belief propagation forbids the sharing
of variables among multiple tables of conditionals in its input,
implying that there must be no loops in the network, but the symmetry
between the posteriors $P(X|Y)$ and the possible observations $P(Y|X)$
allows this dynamic programming algorithm to be recast as a sequence
of ``message updates'' which apply equally well to networks with
variable sharing and therefore loops, as was observed by Pearl in the
1980s \cite{pearl1988pri}. The generalized, message-based form of
belief propagation is called ``loopy belief propagation''
\cite{frey2001very} (it is also called the ``sum-product algorithm''
\cite{kschischang2001factor}), and while this generalization can no
longer be said to compute exact posteriors for its variables, its
numerical behavior and its usefulness in approximation have been the
object of much study, and it enjoys widespread application in certain
specialized domains such as error correcting codes
\cite{wiberg1995codes,mackay1995good}. Loopy belief propagation, to
restate the above definition, is ``exact on trees'', a tree being a
network with no loops, in which case loopy belief propagation reduces
to belief propagation. Its theoretical properties are otherwise
difficult to characterize \cite{ihler2005loopy}, and much research has
been directed towards improving the accuracy of belief propagation by
accounting for the presence of loops in the input model
\cite{jung2006inference,chertkov2006loop,watanabe2008loop,wainwright2003tree,mooij2007lcb}.

\begin{figure}[H]
  \center
  \includegraphics[width=0.8\columnwidth]{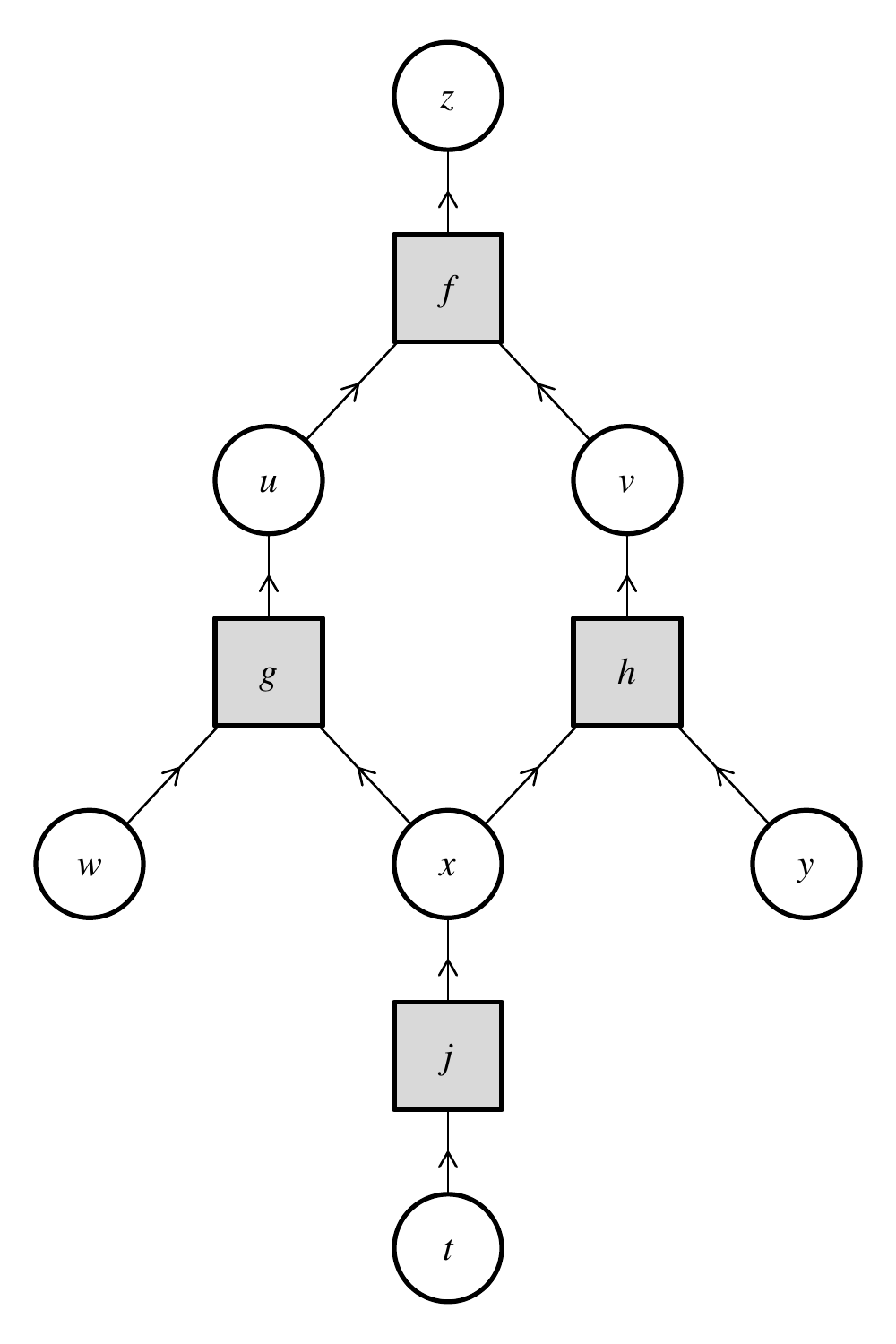}
  \caption{Example network \label{fig:ex-model-bare}}
\end{figure}

This paper considers another class of inputs for which loopy belief
propagation computes exact quantities, namely probabilistic networks
that arise through a straightforward ``lifting''\footnote{ Our
  terminology. For a similar use of the term "lifting" in
  probabilistic inference see \cite{kiselyov2009embedded}; this is
  connected to type-theoretic lifting \cite{hinze2010lifting}; but not
  to be confused with lifted inference
  \cite{poole2003first,desalvobraz2007}, type lifting in compilers
  \cite{saha1998optimal}, or von Neumann's concept of lifting in
  measure theory \cite{tulcea1961lifting}.} of function networks. It
is simple to show that the ``delta function'' posteriors computed by
loopy belief propagation on these networks are exact, as they are just
a probabilistic representation of the deterministic computation
embodied in the original function network. Moreover, we show for the
first time that when the output node of the lifted network is attached
to a Boltzmann distribution
\cite{boltzmann1866mechanische,boltzmann1970weitere} prior, the
messages that propagate backwards through the network encode a
representation of the exact derivatives of the output variable with
respect to each other variable, making loopy belief propagation on the
lifted network an extended or lifted form of backpropagation on the
original function network. This second result is the main contribution
of the paper, establishing that belief propagation is a generalization
of backpropagation.

\section{Example model}

We find it useful to introduce the concepts of this paper through a
small example function network. We define a network containing a
single loop, and a single shared variable $x$, from the following
equations:
\begin{align}
z=f(u,v) \quad u=g(w,x) \quad v=h(x,y) \quad x=j(t)
\end{align}
The network is illustrated in figure \ref{fig:ex-model-bare}.

\section{Backpropagation}

Running backpropagation on this function network means calculating the
derivative of $z$ with respect to the six other variables recursively
using the chain rule of differential calculus, starting with the
variable $u$
\begin{align}
\dby{z}{u}=\partby{z}{u}\equiv\partby{f}{u}\equiv f^{(u)}(u,v)
\end{align}
where we have used ``$\equiv$'' to show the equivalence of alternate
notations for partial derivatives. Then for $v$ we have
\begin{align}
\dby{z}{v}=f^{(v)}(u,v)
\end{align}
and for $w$
\begin{align}
\dby{z}{w}=\dby{z}{u}\partby{u}{w}\equiv\dby{z}{u}g^{(w)}(w,x)
\end{align}
and so on. The term ``adjoint'' is used as a shorthand: since $\ud z$
always appears in the numerator in backpropagation, rather than write
``the derivative of $z$ with respect to $w$'' we call this quantity
``the adjoint of $w$ [with respect to $z$]''
\cite{baydin2018automatic}. Calculating the adjoint of $x$ requires
our first addition:
\begin{align}
\dby{z}{x}=\dby{z}{u}\partby{u}{x}+\dby{z}{v}\partby{v}{x}\equiv\dby{z}{u}g^{(x)}(w,x)+\dby{z}{v}h^{(x)}(x,y)
\end{align}

The last adjoint to be calculated is $t$:
\begin{align}
\dby{z}{t} = \dby{z}{x} j^{(t)}(x)
\end{align}

For a general function network, the chain rule would be written
\cite{lagrange1797,marsden1993basic}
\begin{align}
  \dby{z}{x_i}=\sum_{k \succ i} \dby{z}{x_k}\partby{f_k}{x_i}
\end{align}
where ``$k\succ i$'' means iterating over the parents $k$ of $i$, and
where the general network is defined as a collection of functions and
variables
\begin{align}
x_k = f_k(\{x_i | i\prec k\})
\end{align}

The adjoints of parent variables are calculated and recorded before
their children in a backward pass over the network, giving rise to the
term ``backpropagation'' \cite{rumelhart1986learning}.

\section{Lifting}

We are interested in knowing what happens when we try to run belief
propagation on our network, but first we have to convert the function
network into a probabilistic network with continuous real-valued
variables. To use belief propagation in this setting, we must
represent the variables in our network as probability density
functions. This requires that we first define a probability
distribution over the real numbers which places all of its mass on a
single value:

\begin{align}
P(x=0)=1, \quad P(x\neq 0)=0
\end{align}

\begin{figure}[H]
  \center
  \includegraphics[width=\columnwidth]{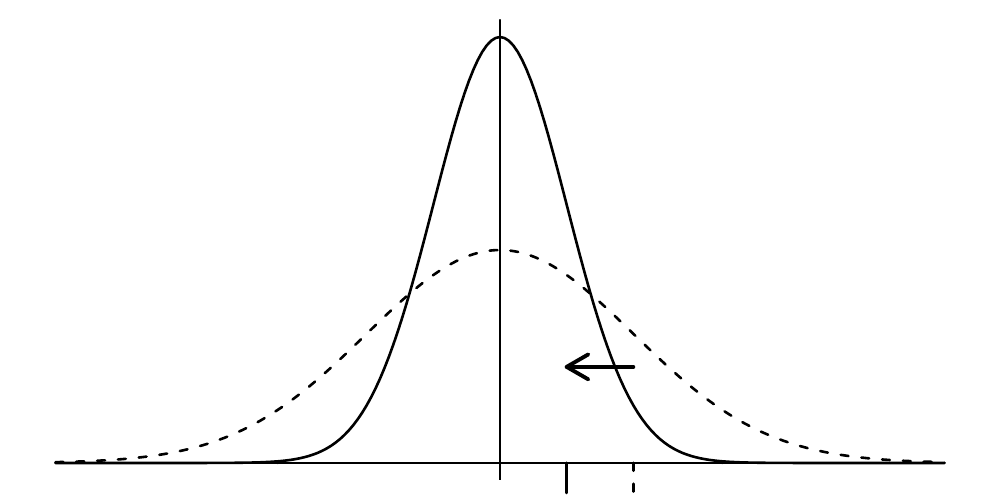}
  \caption{Gaussian distributions getting
    narrower \label{fig:gauss_lim}}
\end{figure}

The density for this distribution is called the ``Dirac delta
function'' \cite{dirac1981principles}, written $\delta(x)$. This is
not a true function since it is infinite at $x=0$, but we can think of
it as a limit of functions, for example a limit of Gaussians whose
standard deviation tends to zero (see figure \ref{fig:gauss_lim}):
\begin{align}
\delta(x) = \lim_{\sigma \to 0} {1 \over \sigma \sqrt{2\pi}} e^{-{1\over 2}({x\over \sigma})^2}
\end{align}
Although this limit itself is not well-defined, it tells us
symbolically how to treat the delta function when it appears inside an
integral, namely by doing the integral first and then taking the
limit:
\begin{align}
\int f(x)\delta(x)\ud x \equiv \lim_{\sigma\to 0} \int f(x) {1 \over \sigma \sqrt{2\pi}} e^{-{1\over 2}({x\over \sigma})^2} \ud x = f(0)
\end{align}
It may be that an algorithmic implementation of our proposed lifting
would approximate delta functions with very narrow Gaussians, in which
case we still expect belief propagation to be well-behaved, but we do
not go into an analysis of that behavior here.

The lifting operation simply replaces each function node $z=f(u,v)$
with a positive-valued ``factor'' defined on all three variables:
\begin{align}
F(u,v,z) = \delta(f(u,v)-z)
\end{align}
which encodes the functional relationship as a density. When working
with such expressions, one must remember that the distinction between
input and output variables hasn't been entirely lost; it is not the
case that $\delta(y-f(x))=\delta(f^{-1}(y)-x)$, because there is a
Jacobian scaling factor:
\begin{align}
\delta(f(x)) &= \abs{f^{(x)}(x)}^{-1}\delta(x), \textrm{\quad hence}  \\
\delta(y-f(x)) &= \abs{f^{(x)}(x)}^{-1}\delta(f^{-1}(y)-x)
\end{align}
%(note that $\abs{f^{(x)}(x)}^{-1}=\abs{(f^{-1})^{(y)}(y)}$).

\section{Belief propagation}

The message updates for (loopy) belief propagation can be written
concisely by defining two types of messages, messages going from
variables to factors, and messages going from factors to variables
\cite{kschischang2001factor}. Messages only go between variables and
the factors to which they are immediately connected; both types of
messages are represented as positive functions of the variable
involved. The message from a variable to a factor is simply the
product of all the messages coming from the other factors to that
variable. For example, referring to figure \ref{fig:ex-model-B}, which shows the lifted
form of the example network, the message from $x$ to $J$ is updated
as:
\begin{align}
m_{(x,J)}(x) := m_{(G,x)}(x) m_{(H,x)}(x)
\end{align}
The message from a factor to a variable is calculated by multiplying
the factor by all of the messages coming into the factor from other
variables, and then integrating (or summing) over the other variables. For example, the message from
$H$ to $x$ is updated as follows:
\begin{align}
m_{(H,x)}(x) := \int H(x,y,v) m_{(v,H)}(v) m_{(y,H)}(y) \ud v \ud y
\end{align}
Convergence of the message updates is usually independent of their
initial values, but for simplicity we assume that they are initialized
to a constant:
\begin{align}
m^0_{(x,J)}(x) := 1 \quad m^0_{(H,x)}(x):=1 \quad \textrm{etc.}
\end{align}
With ordinary (non-loopy) belief propagation, for efficiency the
different messages are updated in a single forward and single backward
pass over the network; any further updates would leave them unchanged,
so they can be said to have converged at this point. Readers who have
encountered Hidden Markov Models
\cite{baum1966statistical,huang1990hidden}, and their continuous,
real-valued counterpart the Kalman filter \cite{kalman1960}, will be
familiar with these forward and backward passes, which are examples of
belief propagation on these specialized probabilistic networks.

With loopy belief propagation, the messages may be updated in any
order. The order of message updates may affect the rate of convergence,
but not the final values to which the messages converge, as long as
convergence is achieved.

After the messages have converged, the posterior of each variable is
estimated as the product of the messages coming into it:
\begin{align}
P(x) \approx {1 \over \int \ud x} m_{(G,x)}(x) m_{(H,x)}(x) m_{(J,x)}(x)
\end{align}
where ${1 \over \int \ud x}$ represents a normalization constant.

\subsection{General form of belief propagation messages}

For reference, we now give the message updates of belief propagation
for a general probabilistic network, consisting of a set of factors
$\{F_\alpha\}$ and variables $\{x_i\}$ (see
\cite{kschischang2001factor}). The message from a factor $F_\alpha$ to
a variable $x_i$ is updated as:
\begin{align}
m_{(F_\alpha, x_i)}(x_i) := \int F_\alpha(x_\alpha) \prod_{j\sim\alpha\-i} m_{(x_j,F_\alpha)}(x_j)\: \ud x_{\alpha\- i} \label{eq:gen.factvar}
\end{align}
where the subscripts $i$ and $j$ index variables in the network, the
subscript $\alpha$ which indexes factors also represents a set of
variables neighboring the respective factor, and $j\sim \alpha\-i$
denotes any variable $j$ neighboring $\alpha$ except $i$.

The update for a message from a variable to a factor is similarly
written:
\begin{align}
m_{(x_i,F_\alpha)}(x_i) := \prod_{\beta\sim i\- \alpha} m_{(F_\beta,x_i)}(x_i) \label{eq:gen.varfact}
\end{align}
\\

\section{Running the lifted model}

In order to simulate evaluating the original function network on a
given set of inputs, we assign priors to the input variables of the
probabilistic network, which we do by attaching single-variable
factors to them. These factors are just delta functions encoding the
input values to the original function network. For example, if the
input values are
\begin{align}
w=w^{*}, \quad t=t^{*}, \quad y=y^{*}
\end{align}
then we introduce factors
\begin{align}
&W^{*}(w)=\delta(w-w^{*}), \quad\nonumber
T^{*}(t)=\delta(t-t^{*}), \\
&Y^{*}(y)=\delta(y-y^{*}) 
\end{align}
These factors will cause messages to propagate upwards through the
network which consist of delta functions that encode the computation
of the original function network.

Finally, we introduce a factor $B$ assigning a Boltzmann prior to the
output node. We omit the arbitrary temperature constant, which can be
recovered by replacing $e$ with $\exp({1/kT})$ in our notation.
\begin{align}
B(z) = e^z
\end{align}

The use of this prior could be seen as representing our desire to
maximize the output of the function network. We will show that it
causes messages to be propagated downward through the network that
make it possible for derivatives to be calculated locally at each
node. The Boltzmann prior is not a true probability distribution,
since it is not normalizable, but this is not a concern for messages.

\begin{figure}[H]
  \center
  \includegraphics[width=0.8\columnwidth]{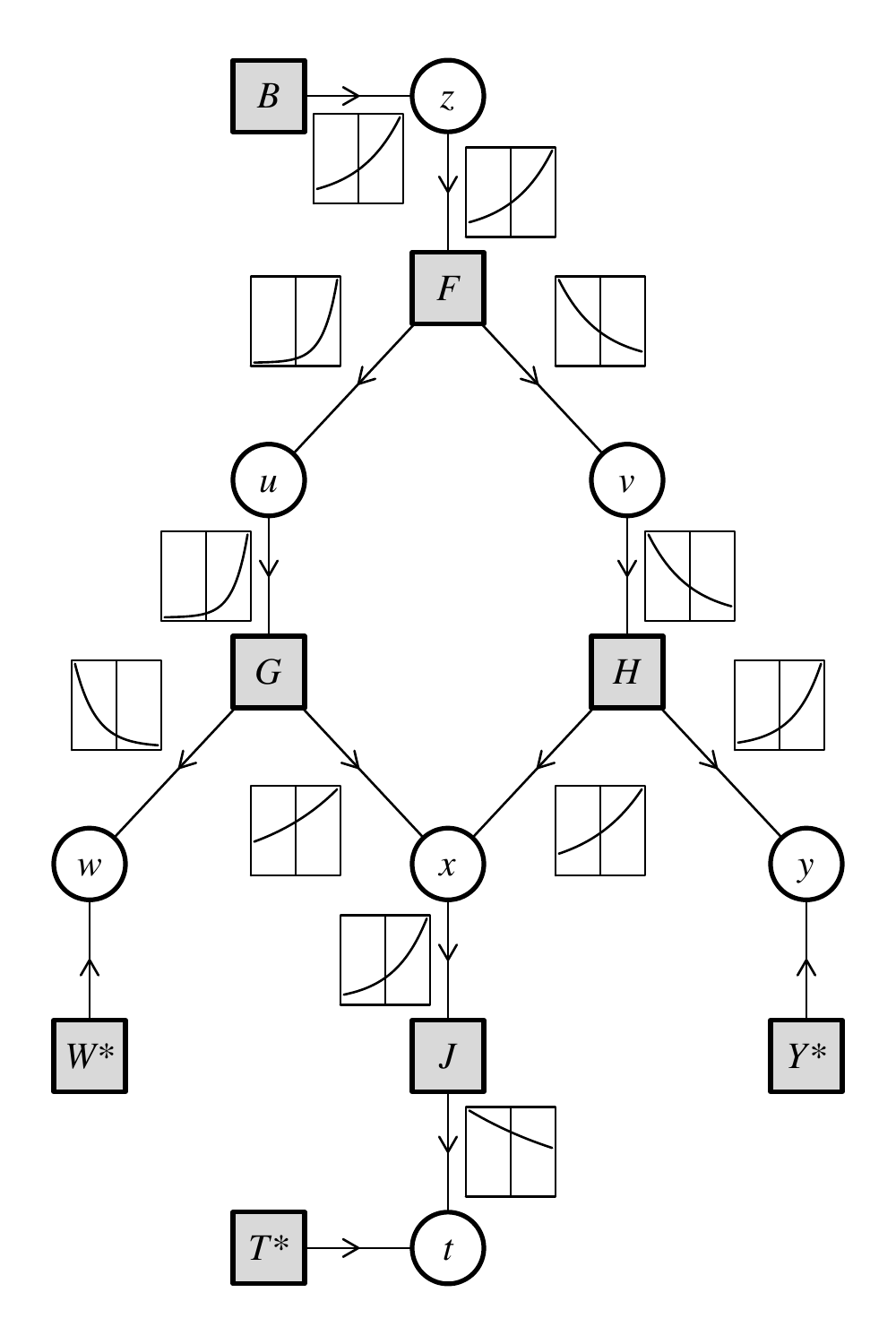}
  \caption{Lifted example network, with messages \label{fig:ex-model-B}}
\end{figure}

The lifted example network is shown in figure \ref{fig:ex-model-B}.
The upward delta function messages have been omitted, but example
downward messages are illustrated with plots next to each edge.

\section{Behavior of the lifted model}

We are now interested in understanding the behavior of belief
propagation on the lifted model. This behavior is specified to a great
extent by the topological structure of the probabilistic network,
which inherits certain properties from the fact that it derives from a
function network. Each factor was originally a function with one or
more input variables and only one output, and with each variable
occurring as the output of at most one function. Therefore although
there is a loop in the network, the network's structure is not
entirely general, and interactions between messages are channeled in
such a way that certain invariants are maintained by the message
updates. In addition to the distinction defined earlier between
variable-factor and factor-variable messages, it is possible to assign
an ``upwards'' or ``downwards'' direction to each message. We find
that when running loopy belief propagation on a network that was
produced by our lifting transformation, messages propagating upwards
through the network are delta functions, but the downward messages can
take an arbitrary form and are not converted into delta functions by
any of the upward messages. When the algorithm converges, the
posteriors at each variable node are delta functions centered at the
value of the variable in the original function network computation,
but the downward messages are nevertheless able to encode additional
information from which the values of derivatives may be obtained.

To see why the upward messages are allowed to take a separate form
from the downward messages, note that each variable has only one
downward message leaving it, towards the factor that it represents the
output of, and that this message is calculated only as the product of
other downward messages coming from factors for which it had served as
an input. Thus, although the product of any function with a delta
function is another delta function, no delta functions enter into the
product when downward messages are updated according to the
variable$\to$factor message updates (equation \ref{eq:gen.varfact}).

The first downward message in the example network comes from the Boltzmann prior $B$:
\begin{align}
m_{(B,z)}(z) = e^z
\end{align}
This is propagated unchanged from $z$ to $F$:
\begin{align}
m_{(z,F)}(z)=e^z
\end{align}
We next calculate the message from $F$ to $U$:
\begin{align}
m_{(F,u)}(u)=\int F(u,v,z) m_{(z,F)}(z) m_{(v,F)}(v) \ud z \ud v
\end{align}
Now $m_{(v,F)}(v)$ is an upward message and therefore a delta
function, $\delta(v-v^*)$. Substituting $F$ according to our lifting,
we have
\begin{flalign}
& m_{(F,u)}(u) & \\
&= \int\delta(z-f(u,v))m_{(z,F)}(z)\delta(v-v^*)\ud v \ud z \label{eq:ex-int-delta} &\\
&= m_{(z,F)}(f(u,v^*)) &\\
&= \exp(f(u,v^*))&
\end{flalign}
This is propagated without change to $G$, the only other neighbor of $u$:
\begin{align}
m_{(u,G)}(u)=\exp(f(u,v^*))
\end{align}
Similarly we have
\begin{align}
m_{(v,H)}(v)=m_{(F,v)}(v)=\exp(f(u^*,v))
\end{align}
Similarly to the message from $F$ to $u$, the message from $G$ to $x$
must incorporate the upward message $m_{(w,G)}(w)$, which is a delta
function $\delta(w-w^*)$: \vspace{0.5em} \\
\begin{flalign}
&m_{(G,x)}(x)&\\
  &= \int\delta(u-g(w,x)) m_{(u,G)}(u) \delta(w-w^*) \ud u \ud w &\\
  &= m_{(u,G)}(g(w^*,x)) &\\
  &= \exp\Bigl(f(g(w^*,x),v^*)\Bigr) &
\end{flalign}
And we see that, correspondingly for $H$,
\begin{align}
m_{(H,x)}(x) = \exp\Bigr(f(u^*,h(x,y^*))\Bigl)
\end{align}
The message from $x$ to $J$ is simply the product of these two
messages:
\begin{align}
m_{(x,J)}(x) = \exp \Bigl(f(g(w^*,x),v^*)+f(u^*,h(x,y^*))\Bigr)
\end{align}
Notice that
\begin{align}
\dby{}{x}\log m_{(x,J)}(x)=\partby{f}{u}\dby{u}{x} + \partby{f}{v}\dby{v}{x} = \dby{f}{x} \label{eq:ex-invar}
\end{align}
when evaluated at $x=x^*$ (and with $w=w^*$ and so on) which gives
$\dby{f}{x}$ according to the chain rule. The relationship only holds
when the derivative is evaluated at $x=x^*$, because $u^*$ and $v^*$
depend on $x^*$, and they appear as constants in the two terms.

It remains to show that the relationship of equation \ref{eq:ex-invar}
holds more generally. Setting aside the example network, let us assume
we are given an arbitrary function network and its lifted counterpart,
a probabilistic network on which we have executed belief propagation.
We want to establish two invariants which hold for the messages in the
network. These invariants apply to downward messages of both types and
relate them to the variable adjoints, which is to say the derivatives
of an objective variable, $z$, with respect to each variable. \\

\begin{thm}
The following invariant holds for the downward message from any
variable $x$ to the factor $F$ adjacently below it:
\begin{align}
\left.\dby{}{x}\right|_{x=x^*} \log m_{(x,F)}(x) = \dby{z}{x} \tag{a}
\end{align}
And the following invariant holds for the downward message from a
factor $F$ with output $y$ to one of its neighboring (input)
variables, $x$:
\begin{align}
\left.\dby{}{x}\right|_{x=x^*} \log m_{(F,x)}(x) = \dby{z}{y}\left(\partby{y}{x}\equiv\partby{f}{x}\right) \tag{b}
\end{align}
\end{thm}

We prove invariants (a) and (b) using induction, by assuming that they
already hold for all the downward-directed messages above the current
edge in the network, and expanding the current message using the
message update rules of equations \ref{eq:gen.factvar} and
\ref{eq:gen.varfact}. After substituting equation \ref{eq:gen.varfact}
into invariant (a), we get
\begin{align}
\dby{}{x}\log\left(m_{(x,F)}(x) = \prod_{G\succ x} m_{(G,x)}(x)\right)
\end{align}
where $G\succ x$ represents any factor $G$ above the variable $x$ in
the network. Since $x$ must be the output node of $F$, this product
iterates over all the neighbors of $x$ not equal to $F$, as specified
by the message update rule. This becomes
\begin{align}
\dby{}{x} \log \prod_{G\succ x} m_{(G,x)}(x) &= \sum_{G\succ x}\left(\dby{}{x} \log m_{(G,x)}(x)\right) \\
  &= \sum_{G\succ x} \dby{z}{g}\partby{g}{x}=\dby{z}{x}
\end{align}
the second equality following from the induction hypothesis and
invariant (b). The lower-case $g$ stands for the function encoded by
the factor $G$ and its output node. This summation is $\dby{z}{x}$ by
the chain rule, which establishes invariant (a) for the message from
$x$ to its child $F$.

To prove the second invariant, we substitute equation
\ref{eq:gen.factvar} into (b), which is to say
\begin{align}
  m_{(F,x)}(x) = \int & \delta\left(\hat{f}-f(x,\{y\})\right) \\
  &m_{(\hat{f},F)}(\hat{f})
  \prod_y m_{(y,F)}(y) \ud\hat{f}\ud\{y\}
\end{align}
where $\hat{f}$ signifies the output variable associated with the
function $f$, and $y$ represents all the inputs of $f$ except $x$. As
with equation \ref{eq:ex-int-delta} above (in our analysis of the
example model), the messages $m_{(y,F)}(y)$ are all delta functions,
so the substitution becomes
\begin{align}
\dby{}{x}\log&\ m_{(F,x)}(x)=\dby{}{x} \log m_{(\hat{f},F)}(f(x,\{y^*\})) \\
&= \left(\dby{}{\hat{f}}\log m_{(\hat{f},F)}(\hat{f})\right)\partby{f}{x}=\dby{z}{f}\partby{f}{x}
\end{align}
where the last equality follows from the induction hypothesis and
invariant (a).

We must finally prove the ``base case'' of the induction, namely that
invariant (a) holds for the message from $z$ to the function node $F$
directly below it. Since the only other neighbor of $z$ is the
Boltzmann factor $B$, this message is equal to the message from $B$ to
$z$:
\begin{align}
m_{(z,F)}(z) = m_{(B,z)}(z) = e^z
\end{align}
Invariant (a) then becomes
\begin{align}
\left.\dby{}{z}\right|_{z=z^*} \log\left(m_{(z,F)}(z)=e^z\right)=\dby{}{z}\log e^z = 1 = \dby{z}{z}
\end{align}
and so it is satisfied for the base case. This completes the proof by
induction.

We have described running belief propagation on a network where the
independent variables are assigned delta function priors:
\begin{align}
X^*(x)=\delta(x-x^*)
\end{align}
In the case where these delta functions are approximated using a more
general distribution such as a narrow Gaussian, it may be more useful
to estimate the adjoints of our variables using a form of invariant
(a) that does not depend on choosing a specific value $x^*$ at which
to evaluate the derivative. \\

\begin{cor}
\begin{align}
\dby{z}{x}=-\int\left(\partby{}{x}X^*(x)\right) \log m_{(x,X^*)}(x)\ud x \label{eq:adj-calc-final}
\end{align}
\end{cor}
For the case of delta functions, this is equivalent to invariant (a)
by analogy to the following integration by parts identity:
\begin{align}
\int \left(\partby{}{x}\delta(x)\right) f(x)\ud x = -\partby{f}{x}(0)
\end{align}
But when $X^*$ is a Gaussian, for example, the above expression
\ref{eq:adj-calc-final} for $\dby{z}{x}$ is equivalent to a kind of
smoothed numerical differentiation.

Our proof by induction makes it clear that as with backpropagation,
the converged belief propagation messages in our model can be
calculated in a single forward and single backward pass over the
network.

\section{Conclusions}

\subsection{Motivation}

Most papers in machine learning seek to introduce a new computer
algorithm to the field. The purpose of this paper is rather to shed
light on a connection between two well-established algorithms, to
provide groundwork for a better theoretical understanding of both
algorithms, and to eliminate some of the mystery surrounding them for
students.

Belief propagation and backpropagation apply to different classes of
input model. Belief propagation applies to probabilistic models and is
used in domains where there is a need to model uncertainty directly,
and backpropagation applies to deterministic models, where it is used
to provide gradients to support the fitting of such models to data.
Because all real-world data contains some measure of uncertainty,
there is considerable overlap between these two domains, and it could
be said that any essential difference between them is only a matter of
engineering philosophy; based on the engineer's decision about whether
to model uncertainty directly or indirectly, and at which level of the
system to do so \cite{jaynes2003probability,murphy2012machine}.

There has been recent interest in extensions of backpropagation that
incorporate uncertainty more directly into the algorithm; some of
these, such as stochastic gradient descent \cite{kiefer1952stochastic}
or drop-out \cite{srivastava14a}, apply backpropagation to inputs
which change at random; others, such as probabilistic backpropagation
\cite{pmlr-v37-hernandez-lobatoc15}, extend backpropagation by
replacing deterministic quantities with probabilistic representations
of the same quantities, somewhat related to the ``lifting'' we refer
to in this paper. We hope that it would be possible to assist these
investigations by clarifying the mathematical relationship between
backpropagation and belief propagation.

\begin{figure}[H]
  \center
  \includegraphics[width=0.7\columnwidth]{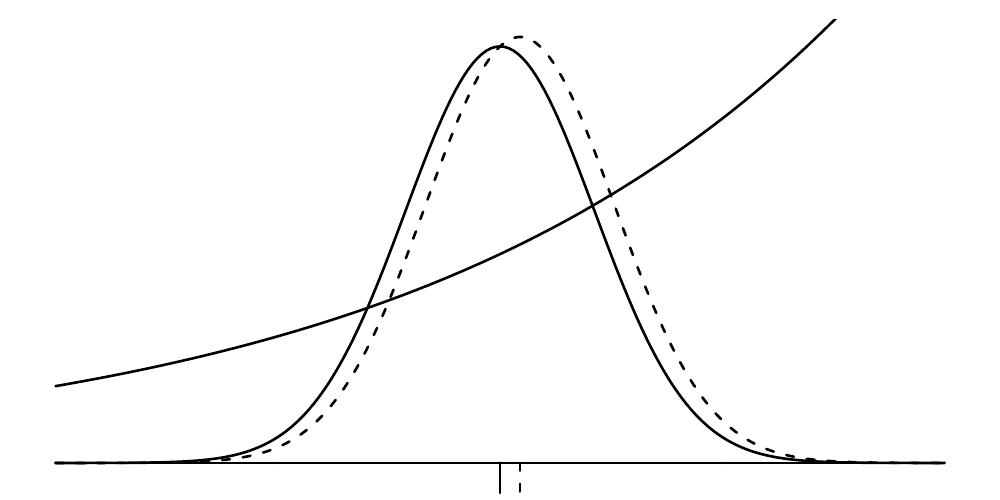}
  \caption{A Gaussian message being shifted to the right after
    multiplication by an exponential \label{fig:gauss_boltz}}
\end{figure}

The problem of characterizing the behavior of belief propagation on a
lifted function network whose inputs have been initialized with
distributions other than delta functions remains an open question. In
this case, we can expect in general that the converged messages will
not produce exact posteriors and will not lead to exact adjoints being
calculated, because the downward messages will have the effect of
slightly changing the variable locations specified in the upward
messages (figure \ref{fig:gauss_boltz}) and these effects will be
compounded as the messages propagate around loops. We do not know
whether a loop-corrected form of belief propagation would be necessary
to make this more general scenario useful.

However, before learning the inputs to a function network using
gradient descent, the precise value of the input variables is in
general unknown. Being able to make this uncertainty explicit at a
more basic level, by running some form of backpropagation on a
``lifted'' probabilistic version of the network, with inputs that are
not delta functions, could be desirable for a number of reasons, for
example because it allows the convergence rate of the input variables
to be reflected in their posterior distributions, or because it allows
some of the input variables to be specified with less certainty than
others, which could provide an evolving indicator of where the
training algorithm should focus its attention.

Readers who are interested in probabilistic approaches to the problem
of training ``neural networks'' could start with David MacKay's thesis
\cite{mackay1992bayesian} which proposes approximations that could be
used to model uncertainty at the level of variables in the network.
Extensions to this idea are explored in for example
\cite{solla1998optimal} and \cite{braunstein2006learning}; more
recently, \cite{soudry2014expectation} points out that by replacing
backpropagation with message passing, it becomes easier to train
networks that have discrete weights, which can be useful for
hardware-based network implementations with limited numerical
precision. ``Probabilistic backpropagation''
\cite{pmlr-v37-hernandez-lobatoc15} is the name given to an approach
that combines a forward pass that approximates the distribution at
each network node as a Gaussian, with a backwards pass that
backpropagates adjoints of these distribution parameters. Experiments
show that the method compares favorably with plain backpropagation
and with Hamiltonian Monte-Carlo, a probabilistic training method
based on sampling \cite{neal1995bayesian}, although there appear to be
many details in the implementation. Our paper is less concerned with
experimental results, and more concerned with making a sea of
different ideas more navigable by pointing out some overlooked
connections that exist within it.

Belief propagation and backpropagation are both useful for analyzing
large models because they have the same time complexity as running the
model itself. Like strict (non-loopy) belief propagation,
backpropagation is a dynamic programming algorithm that requires only
two passes over the input network, the first pass serving to compute
the value of the objective or output variable, and the second pass
serving to compute the derivatives. Loopy belief propagation, on the
other hand, exchanges numerical messages locally on the network for
some usually small number of iterations, typically until the messages
converge; for error-correcting codes and certain other applications,
convergence is fast enough that the algorithm does not add significant
time complexity
\cite{mackay1995good,wiberg1995codes,richardson2000geometry,murphy2013loopy}.
While belief propagation and backpropagation both distinguish between
input and output variables, loopy belief propagation requires no such
distinction to be made.

Framing an algorithm in terms of locally-exchanged messages can be
useful for distributing it across multiple computers, and there may be
some value derived from being able to rethink backpropagation in terms
of iterative local message-passing. Another contribution of this paper
is to show that by placing backpropagation in the framework of loopy
belief propagation, the input-output relationships of backpropagation
become part of the messages rather than being hard-coded through the
functions of the network, and the original function network can be
inverted with respect to one of the input variables, simply by moving
the Boltzmann prior onto this variable while leaving the rest of the
network unchanged.

\subsection{Generality}

It is desirable to point out that the form of the ``lifting'' of a
function network to a probabilistic network which we describe here is
a straightforward requirement of the problem of converting from one
class of inputs to the other. The ``Dirac delta function'' is a
well-understood formalism for specifying a probability distribution
that takes only a single value, and it is used to lift both variables
and functions into the domain of probabilities.

The use of the Boltzmann distribution is motivated as follows:
backpropagation is most commonly used to solve optimization problems;
the most natural way of converting an optimization problem to a
probabilistic inference problem is to place a Boltzmann distribution
over the objective: $p(E) \propto \exp{E\over kT}$, where $E$ is the
objective or output variable of the function network, and $kT$ is a
constant specifying the tightness of the distribution around the
optimum. This distribution has its origins in thermodynamics, where it
describes the distribution over the states of a system with energy $E$
and temperature $T$ \cite{fermi1936}. Also called the Gibbs measure in
mathematical contexts, the Boltzmann distribution has widespread use
in machine learning, for example in stochastic neural networks, see
for example the ``Boltzmann machine''
\cite{sherrington1975,ackley1985}; and arises almost universally in
probability theory in a less recognizable form, the exponential family
model, which appears whenever data consist of exchangeable
observations \cite{wainwright2003graphical}.

There are a few hurdles to overcome in attempting to unify belief
propagation and backpropagation. The first is that the domains of each
algorithm are different, one being probabilistic and the other being
deterministic. This is addressed by our ``lifting'' transformation,
but the transformation produces models that are considered less
tractable than a typical input to belief propagation: first of all, a
typical lifted function network will contain many loops; and secondly,
a function network operates on real-valued rather than discrete
variables. Computing the message updates of belief propagation on the
lifted network requires some difficult modeling decisions: how to
represent distributions over real variables, whether to represent
delta functions specially or as a limit of narrow Gaussians, how to
perform the numerical integration required by the message updates, and
how to represent the Boltzmann distribution and other messages which
may be unnormalizable. All of these hurdles can be surmounted in
various ways. There is a relatively long history of the successful use
of belief propagation and related message-passing algorithms to
perform efficient probabilistic inference in real-valued probability
networks, see for example ``assumed density filtering'' and
``expectation propagation'' \cite{minka2001}.

Finally, this work has relevance to researchers seeking to invent
novel ways to improve the training phase of models based on function
networks, which are seeing increasingly widespread application in
computer science. Rather than changing the structure or mathematical
relationships of the network to make it behave more tractably under a
backpropagation-based training method, one could instead consider
tuning its independent variables by applying belief propagation (or
one of its many extensions) to a lifted, probabilistic, version of the
network where it is possible to reason about uncertainty more
directly.

To this end, it would seem helpful to observe that the original
backpropagation algorithm is recovered exactly by loopy belief
propagation in the case where the network is initialized with delta
functions, just as it has often been helpful in the analysis of loopy
belief propagation on general probabilistic networks to observe that
the algorithm is exact in the case where the input network is a tree.

\bibliographystyle{unsrt}
\bibliography{bibdata}

\end{document}